\documentclass{article}
\usepackage[dblblindworkshop, final]{neurips_2026}
\workshoptitle{The 6th Workshop on Mathematical Reasoning and AI}
\usepackage[utf8]{inputenc}
\usepackage[T1]{fontenc}
\usepackage{hyperref}
\hypersetup{hidelinks}
\usepackage{url}
\usepackage{booktabs}
\usepackage{amsmath,amsfonts,amssymb}
\usepackage{microtype}
\usepackage{graphicx}
\usepackage{verbatim}
\title{Can an AI Agent Rediscover a Blaschke-Curve Invariant?}
\author{Yunus Zeytuncu\\
University of Michigan-Dearborn\\
\texttt{zeytuncu@umich.edu}}
\newcommand{\T}{\mathbb{T}}
\begin{document}
\raggedbottom
\maketitle

\begin{abstract}
We study generalized Blaschke curves as a controlled environment for AI-assisted mathematical rediscovery. For one fixed degree-four Blaschke product, an agent receives numerical coordinates of the six pair-lines determined by each of 80 boundary configurations. The target theorem is withheld from the task instructions. The saved research log reports rejected geometric hypotheses and a homogeneous cubic fitted to polygon sides. Its frozen coefficients predict 480 lines from 80 unseen parameter values, with a recorded RMS scale-free residual of $8.88\times10^{-17}$. Discovery-set diagonals provide an out-of-fit consistency check, not a fully held-out test. A separate one-configuration run reports insufficient evidence for invariance. A post-review deterministic degree-search baseline also recovers the cubic, so the experiment does not establish an advantage over polynomial fitting. We present this single-instance case study as a protocol for separating conjecture, numerical validation, and proof, with explicit limitations concerning agent metadata, prior knowledge, and reproducibility.
\end{abstract}

\section{Introduction}
An experimentally suggested equation is not yet a theorem, and a successful numerical fit is not by itself evidence of an autonomous discovery process. These distinctions motivate a small, inspectable test case in Blaschke geometry. We ask whether a mathematical agent can formulate a parameter-independent relation from numerical observations and commit to that relation before testing new configurations.

Scientific rediscovery from data has an established history. For example, AI Feynman evaluates symbolic regression on known physical equations \citep{udrescu2020}. FunSearch and AlphaGeometry illustrate different combinations of learned search, mathematical structure, and verification \citep{romera2024funsearch,trinh2024alphageometry}. Our contribution is narrower: a geometric case study with an explicit information boundary and a frozen numerical prediction. We do not introduce symbolic regression, prove a new Blaschke theorem, or estimate an agent success rate.

Finite Blaschke products connect complex analysis, Poncelet geometry, and numerical ranges \citep{daepp2019,mirman2005}. In the normalized degree-three case, the roots of a boundary level equation form triangles tangent to a fixed ellipse. Higher degrees lead to generalized algebraic envelopes \citep{siebeck1864,linfield1920,hunziker2022}. This gives a rediscovery target that can be hidden from task instructions while remaining mathematically grounded. The resulting experiment separates three questions: whether a candidate equation fits, whether it predicts new parameter values, and what the available record establishes about the agent's process.

\section{A geometric discovery environment}
Write $\T=\{z\in\mathbb C:|z|=1\}$ and fix
\begin{equation}
B(z)=z\prod_{j=1}^{3}\frac{z-a_j}{1-\overline{a_j}z},\qquad
(a_1,a_2,a_3)=(0.22+0.18i,-0.31+0.09i,0.08-0.38i).
\label{eq:blaschke}
\end{equation}
For each $\lambda\in\T$, the four roots of $B(z)=\lambda$ lie on $\T$. In cyclic order, they determine four sides and two diagonals. We encode each pair-line by $[u:v:w]$, where $ux+vy+w=0$, using $u^2+v^2=1$ and a deterministic sign. A point in this dual projective plane represents a line in the original plane.

The known geometric structure is a fixed cubic relation among these line coordinates. Equivalently, the pair-lines are tangent to a parameter-independent algebraic envelope of class three; the class refers to the degree of its dual curve, not necessarily its degree in the original plane. The experiment seeks numerical evidence for this relation without supplying the theorem or its expected degree. It does not seek the zeros from raw observations: both the representation and the data-generation problem were chosen by the experimenter.

\begin{figure}[t]
\centering
\includegraphics[width=.45\linewidth]{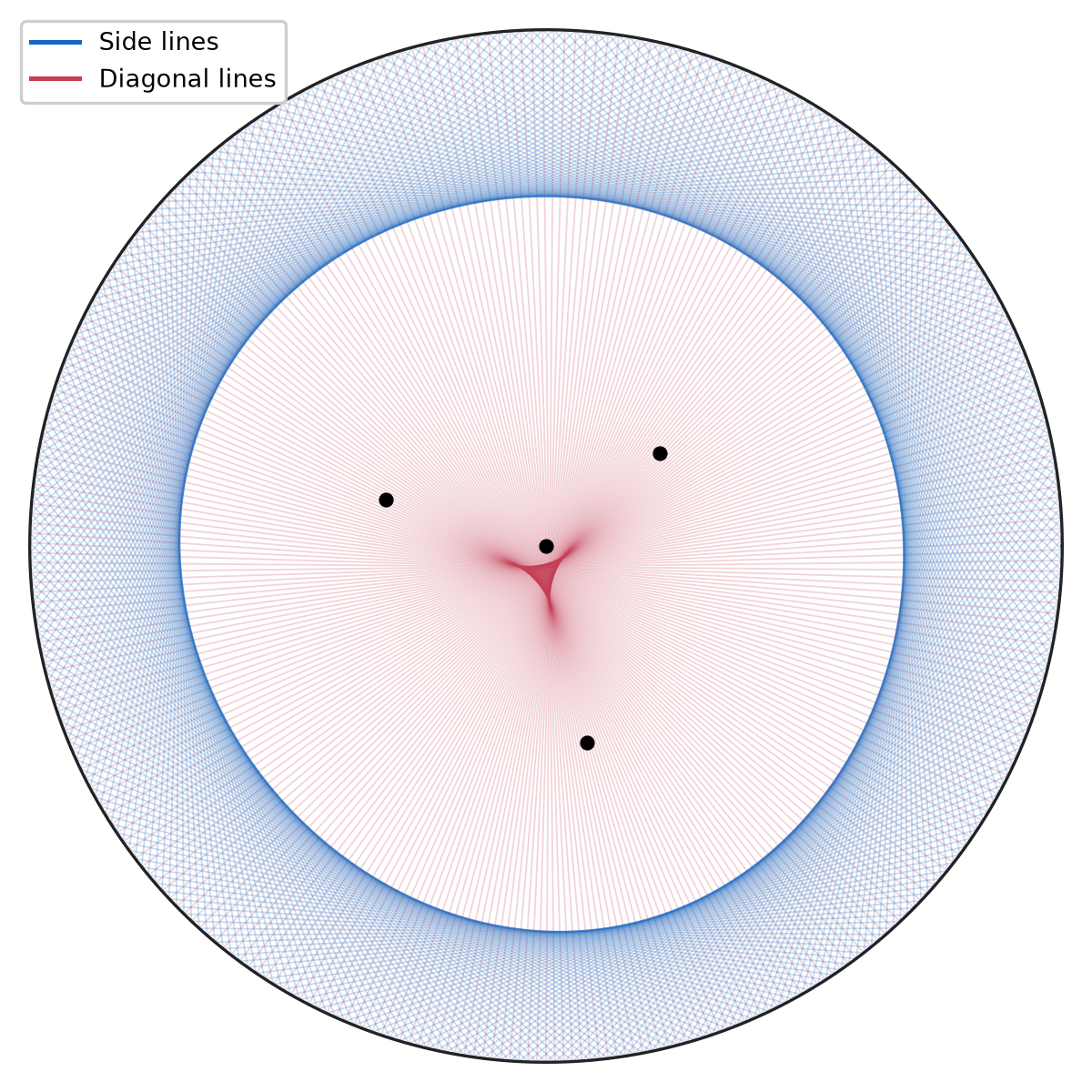}
\caption{The 480 stored discovery lines from exactly 80 values of $\lambda$. Blue lines are sides and red lines are diagonals; black points are the zeros of $B$. This post-review visualization uses the experimental discovery CSV, not an additional sampling grid. No envelope figure was supplied in the original discovery task.}
\label{fig:envelope}
\end{figure}

\paragraph{Data construction.}
For $j=0,\ldots,159$, let $\theta_j=2\pi(j+0.173)/160$. Even indices provide discovery data and odd indices provide evaluation data. Each split contains 80 configurations and 480 lines. We solve
\[
z\prod_{j=1}^3(z-a_j)-e^{i\theta}\prod_{j=1}^3(1-\overline{a_j}z)=0
\]
numerically, project each root radially onto $\T$, sort by argument, and compute the six pair-lines. The stored modulus diagnostic is measured \emph{after} projection and is not an independent root-accuracy estimate. The split tests new values of the same parameter for the same product, not new products or extrapolation beyond the sampled circle.

\section{Agent protocol and recorded trajectory}
\paragraph{Information boundary and audit.}
The archived protocol describes a separate Codex sub-agent receiving the discovery CSV and a task prompt. Recovered local session records identify both agents as \texttt{gpt-5.6-sol} with \texttt{low} reasoning effort. Both launches specify \texttt{fork\_turns: none}, excluding a fork of the parent conversation. The CSV includes parameter values, side/diagonal labels, and line coordinates; the prompt names degree-four Blaschke products and asks for stable algebraic or geometric structure. Numerical or symbolic calculations, fitting, and plotting are allowed. The theorem, expected degree, held-out file, other workspace sources, external lookup, and parent-agent hints are prohibited during discovery. File restrictions are instruction-level: the recorded execution policy permits unrestricted filesystem access.

The protocol states that the parent agent audited the written conjecture before releasing the held-out file, with coefficients and thresholds then fixed. The human experimenter selected the mathematical setting; the computational scaffold generated data and managed release. The recovered parent record contains two experiment launches and an evaluation follow-up for each. Child records preserve tool calls and usage metadata, but a complete launch-message and intervention audit remains unavailable. Temperature, a sampling seed, a dated model snapshot, and a prespecified execution budget are not established. We report two documented trajectories without claiming an exhaustive account-wide search for attempts. Appendix~\ref{app:records} gives recovered metadata and remaining limitations.

\paragraph{Reported hypothesis development.}
The full-data log first rejects a fixed intersection point for the two diagonals. It then fits a homogeneous quadratic to the 320 side lines, obtaining an algebraic RMS residual of $9.25\times10^{-3}$ in its chosen coefficient scaling. Splitting sides by the sign of $w$ reduces but does not eliminate this error. These checks reject those particular descriptions, not every conceivable conic decomposition.

The log next reports a search over homogeneous polynomial degrees. At degree three, the side-only design matrix has smallest and next-smallest singular values $3.21\times10^{-15}$ and $6.64\times10^{-2}$. Four interlaced discovery-parameter folds yield small omitted-fold residuals. The 160 discovery diagonals also satisfy the side-fitted cubic, with raw RMS residual $8.29\times10^{-17}$. Since those diagonals were visible from the outset, this is an \emph{out-of-fit consistency check}, not a hidden-label generalization test.

\paragraph{Frozen numerical conjecture.}
The monomials are ordered as
\[
(u^3,u^2v,u^2w,uv^2,uvw,uw^2,v^3,v^2w,vw^2,w^3).
\]
The archived equation is
\begin{align}
P(u,v,w)={}&0.0032683124u^3-0.0494655836u^2v+0.491767409872u^2w\nonumber\\
&-0.0171636876uv^2-0.0198uvw+0.01uw^2-0.0202735836v^3\nonumber\\
&+0.502767409872v^2w+0.11vw^2-w^3=0.
\label{eq:cubic}
\end{align}
The agent conjectures this relation for every pair-line of the fixed product. Its prespecified test uses
\begin{equation}
r(u,v,w)=\frac{|P(u,v,w)|}{(u^2+v^2+w^2)^{3/2}},
\label{eq:residual}
\end{equation}
requiring every evaluation row below $10^{-10}$ and RMS below $10^{-11}$, both overall and separately for sides and diagonals. This residual is invariant under rescaling the \emph{line} coordinates. Its value still depends on polynomial scaling, fixed here by the coefficient $-1$ of $w^3$.

\section{Results and deterministic controls}
\paragraph{Original frozen evaluation.}
The archived report records 480/480 passing rows, overall RMS $8.88\times10^{-17}$, and maximum $3.18\times10^{-16}$. Side and diagonal RMS values are $8.86\times10^{-17}$ and $8.94\times10^{-17}$, respectively. These are numerical predictions at previously unseen parameter values, not proof of an identity for all $\lambda$. A post-review reevaluation of the stored coefficients and CSV gives RMS $8.85\times10^{-17}$; the tiny difference is at floating-point roundoff scale. We retain the original recorded result rather than replace its history.

\paragraph{One-configuration ablation.}
The separate ablation retained \emph{only one} discovery parameter value and its six lines; it did not remove just one configuration from the full set. The saved log diagnoses inadequate evidence for parameter invariance, records unit-circle and incidence relations, and produces no invariant cubic. Its construction-level check passes the 80 evaluation configurations but is not the target discovery. With ten homogeneous cubic coefficients and at most six independent linear constraints, one configuration cannot uniquely determine a cubic up to scale from incidence data alone. This dimension count does not rule out identification under additional prior assumptions.

The ablation changes both the number of parameter values and the number of observations. Its contrast with the successful trajectory therefore does not isolate a causal effect of parameter diversity, establish that every one-configuration prompt fails, or quantify reliability across agents.

\paragraph{Post-review SVD baseline.}
We added a deterministic comparison after review; it was not part of the original agent protocol. For each degree $d=1,\ldots,5$, we form the matrix of all degree-$d$ monomials evaluated at the same 320 discovery side rows. We count singular values above $10^{-12}\sigma_{\max}$ and select the first degree with a one-dimensional numerical nullspace. Degrees one through five have nullities $0,0,1,3,6$. The higher-degree relations are consistent with multiples of the cubic. No diagonal or evaluation row enters this degree selection or coefficient fit.

\begin{table}[t]
\centering
\caption{Original agent outcome and post-review deterministic comparison. Both polynomials use coefficient $-1$ for $w^3$ and residual~\eqref{eq:residual}. Roundoff-level differences are not evidence of superior discovery performance.}
\label{tab:results}
\begin{tabular}{lrr}
\toprule
Procedure & Evaluation RMS & Evaluation maximum\\
\midrule
Frozen agent cubic (archived report) & $8.88\times10^{-17}$ & $3.18\times10^{-16}$\\
Deterministic degree search (new control) & $1.24\times10^{-15}$ & $3.18\times10^{-15}$\\
\bottomrule
\end{tabular}
\end{table}

The baseline selects degree three and passes the same numerical thresholds (Table~\ref{tab:results}). After unit-norm scaling and sign alignment, its coefficient vector differs from the frozen vector by $6.20\times10^{-15}$ in Euclidean norm. Thus the experiment does not demonstrate that an agent is needed to recover the cubic from this representation. Its process-level interest lies in formulating and recording hypotheses and validation criteria, not in outperforming SVD; these aspects require stronger evaluation in a larger study.

\paragraph{Parameter-count diagnostic.}
As a further numerical control, we select one, two, five, ten, twenty, forty, or eighty evenly spread discovery configurations and repeat their side rows to give 320 rows in every condition. The cubic nullities are $6,2,1,1,1,1,1$, respectively. Repetition changes row count but not information or rank. For each tested $k\geq5$, the fitted cubic predicts evaluation lines with unit-coefficient RMS below $2.3\times10^{-15}$. These deterministic diagnostics show why replicated observations cannot repair insufficient independent constraints. They are not repeated agent runs, a matched-information experiment, or a claim that five configurations are minimal. Appendix~\ref{app:baseline} specifies the computation.

\section{Interpretation and limitations}
The study supports a modest conclusion: a documented agent trajectory produced a numerical equation that predicts new configurations of one fixed Blaschke product. Parameter-separated evaluation avoids training and testing on different lines of the same configuration, and freezing the equation makes the numerical claim directly checkable. Nevertheless, held-out parameters alone do not establish novelty, proof, or a uniquely agentic mechanism.

Withholding a theorem from the prompt is not equivalent to withholding it from pretraining. The task explicitly names Blaschke products, and the agent could have relevant prior knowledge. Research logs record reported hypotheses; they cannot certify faithful internal reasoning or rule out recalled results. This distinction is supported by studies of explanation faithfulness \citep{turpin2023,chen2025}. We therefore use rediscovery in the operational sense of producing and testing a withheld target relation, not as a claim about the origin of the model's knowledge.

The principal limitations are a single product, one documented trajectory per condition, unrecorded sampling settings and execution budgets, and incomplete process auditing. Numerical rank deficiency is not a symbolic proof. The mathematically informed dual-coordinate representation also makes deterministic recovery straightforward. Future experiments should prerecord complete configurations and budgets, compare repeated runs against numerical baselines, hide diagonals entirely during discovery, vary products and representations, and distinguish symbolic verification from numerical accuracy. The present case study is a starting point for such a workflow, not its completed validation.

\clearpage
\begin{ack}
The author has no grant funding to acknowledge and declares no competing interests.

AI tools assisted the experiment, analysis, and preparation of this manuscript. The author is responsible for checking the mathematics, reported evidence, citations, and final text. The post-review controls are identified separately from the archived agent runs.
\end{ack}

{\small
\bibliographystyle{plainnat}
\bibliography{references,references_camera_ready}
}
\clearpage
\appendix
\section{Available records and reproducibility boundaries}
\label{app:records}
\paragraph{Code and data availability.}
The code, numerical data, saved task prompts and research logs, and recovered run metadata are publicly available in the following repository:
\begin{center}
\url{https://github.com/yezeytuncu/blaschke-ai-rediscovery}.
\end{center}
The experimental artifacts used here are preserved in repository commit \href{https://github.com/yezeytuncu/blaschke-ai-rediscovery/tree/578b7840744874426c8809759509c6969598a662}{\texttt{578b784}}. Raw application conversations are not included.

The project records contain the generator, discovery and ablation CSV files, task prompts, and separate protocol, research-log, and evaluation files. On September 29, 2026, we additionally recovered metadata and tool-call records from the original local application sessions dated August 26. The prompts below are the saved task prompts, not a reconstruction of the full system prompt or complete conversation. The original summaries remain unchanged; a separate metadata audit records the recovered information without publishing raw app logs.

Both child sessions record provider \texttt{openai}, model identifier \texttt{gpt-5.6-sol}, reasoning effort \texttt{low}, and harness version \texttt{0.150.0-alpha.8}. The model identifier is not a dated weights snapshot. Parent launch calls specify \texttt{fork\_turns: none} and no explicit model override. The full-data run used shell commands and R scripts after a failed Python/NumPy import; the sparse-data run used Python standard-library calculations. Both wrote their reports through file-editing tools. The visible child tool calls introduce the evaluation CSV only in the second task turn. This supports the recorded discovery/evaluation ordering but is not a complete audit of launch-message content or all possible context exposure.

\begin{table}[h]
\centering
\caption{Recovered original-run accounting. Elapsed times span recorded task start to completion. Output-token counts are the harness-reported field, not a prespecified budget or a cost estimate.}
\begin{tabular}{llrr}
\toprule
Run & Phase & Elapsed seconds & Output tokens\\
\midrule
Full data & Discovery & 143.1 & 6,219\\
Full data & Evaluation & 43.3 & 2,129\\
Sparse data & Discovery & 122.7 & 5,209\\
Sparse data & Evaluation & 100.8 & 4,815\\
\bottomrule
\end{tabular}
\end{table}

The audit exports only selected metadata, timestamps, usage counters, launch options, source line numbers, and source-file hashes. It excludes raw messages and reasoning content. Temperature, random seed, original R/Python versions, and a prespecified execution budget remain unestablished. The inspected parent session contains two experiment launches; this is not an exhaustive account-wide attempt count or evidence of a prospective run-selection rule. These limitations prevent exact agent-level reproduction even though the numerical computations are independently checkable.

\paragraph{Data and numerical computation.}
The original generator is named \texttt{blaschke\_invariant.py} in the experiments directory. The CSV files are stored in \path{outputs/degree4_baseline/}. Each row contains \texttt{theta}, the real and imaginary parts of $\lambda$, \texttt{chord\_type}, and $(u,v,w)$. The sign convention makes the first line coefficient with magnitude above $10^{-12}$ positive after normalization by $\sqrt{u^2+v^2}$. This convention fixes a CSV representation, not an additional geometric condition.

The generator itself includes a cubic-fitting calculation. This experimenter-side computation is not evidence of agent discovery and was not an allowed discovery input under the archived protocol. The relevant agent outcome is the separately recorded frozen polynomial and its evaluation. The discovery and evaluation CSV files remain unchanged in the camera-ready preparation.

\subsection*{Saved task prompt for the full-data trajectory}
\begingroup\small
\begin{minipage}{\linewidth}
\verbatiminput{agent_prompt.txt}
\end{minipage}
\endgroup

\subsection*{Saved task prompt for the sparse-data trajectory}
\begingroup\small
\begin{minipage}{\linewidth}
\verbatiminput{ablation_prompt.txt}
\end{minipage}
\endgroup

\paragraph{Protocol scaffold.}
The protocol permits only the respective prompt and discovery CSV before freezing; it prohibits evaluation data, other workspace files, browsing, external theorem lookup, and parent-agent hints. The full-data agent must record discarded hypotheses, a precise conjecture, coefficients, and thresholds. The parent agent then audits the written log and releases only the evaluation CSV, with no refitting allowed. These are the recorded instructions and phase boundaries. We do not infer a hardware-enforced isolation mechanism or a faithful hidden reasoning trace from them.

\section{Post-review baseline and parameter-count controls}
\label{app:baseline}
The new script \texttt{experiments/camera\_ready\_baseline.py} reads the stored CSV files without regenerating them. Its monomial order enumerates exponents $(i,j,d-i-j)$ with $i$ decreasing from $d$ to zero and, for each $i$, $j$ decreasing from $d-i$ to zero. At $d=3$, this is precisely the order used in equation~\eqref{eq:cubic}. Singular-value decomposition is performed on the raw monomial matrix, without column standardization. The relative numerical rank threshold is $10^{-12}$. If there are fewer rows than columns, nullity includes the additional right-nullspace dimensions. Coefficients are initially normalized to Euclidean norm one. For Table~\ref{tab:results}, the cubic is rescaled so that its $w^3$ coefficient is $-1$.

\begin{table}[h]
\centering
\caption{Numerical degree search on the 320 discovery side rows.}
\begin{tabular}{rrrr}
\toprule
Degree & Number of monomials & Numerical nullity & $\sigma_{\min}/\sigma_{\max}$\\
\midrule
1 & 3 & 0 & $9.46\times10^{-1}$\\
2 & 6 & 0 & $8.71\times10^{-3}$\\
3 & 10 & 1 & $2.75\times10^{-16}$\\
4 & 15 & 3 & $1.37\times10^{-16}$\\
5 & 21 & 6 & $8.17\times10^{-17}$\\
\bottomrule
\end{tabular}
\end{table}

For the parameter-count diagnostic, sort the 80 discovery angles and select indices $\lfloor80j/k\rfloor$, $j=0,\ldots,k-1$. Each selected configuration supplies its four side lines, and each selected row is repeated $80/k$ times. All tested $k$ divide 80. This creates 320 rows in every condition. The numerical nullity is checked against that of the unreplicated subset; the two agree in every reported case. Evaluation is reported only for a one-dimensional cubic nullspace, since an arbitrary vector from a larger nullspace is not a uniquely identified equation.

\begin{table}[h]
\centering
\caption{Post-review deterministic controls, not additional agent ablations. Residuals here use unit Euclidean coefficient norm, unlike Table~\ref{tab:results}.}
\begin{tabular}{rrrr}
\toprule
Configurations & Unique side rows & Cubic nullity & Evaluation RMS\\
\midrule
1 & 4 & 6 & Not uniquely identified\\
2 & 8 & 2 & Not uniquely identified\\
5 & 20 & 1 & $2.00\times10^{-15}$\\
10 & 40 & 1 & $2.22\times10^{-15}$\\
20 & 80 & 1 & $2.00\times10^{-15}$\\
40 & 160 & 1 & $7.78\times10^{-16}$\\
80 & 320 & 1 & $1.01\times10^{-15}$\\
\bottomrule
\end{tabular}
\end{table}

The controls were run with Python 3.12.4, NumPy 2.5.3, and Matplotlib 3.11.2 on macOS arm64. They use no random sampling. The output JSON records software versions, input SHA-256 hashes, singular-value diagnostics, coefficient comparisons, residuals, and repeated-row controls. These are versions for the new computations, not recovered versions for the original agent runs. Floating-point summation order and numerical libraries can change the final few digits at roundoff scale.

The baseline was designed with knowledge of the original result. Although its code fits and selects using discovery sides alone, this is a retrospective comparison, not a newly blinded experiment. Its role is to determine whether a simple non-agent procedure suffices on the published instance. Repeated rows are deliberately redundant and cannot substitute for a future study of distinct observations, independent product instances, or repeated agent trials.
\end{document}